\documentclass[11pt]{article}
\usepackage{bm} 
\usepackage{graphicx}
\usepackage{epstopdf}
\usepackage{float}
\usepackage{amsthm}
\usepackage{natbib}
\usepackage{amsmath}
\usepackage{float}

\newtheorem{theorem}{Theorem}
\makeatletter

\begin{document}
\title{Any Target Function Exists in a Neighborhood of Any Sufficiently Wide Random Network: A Geometrical Perspective}
\author{Shun-ichi Amari \thanks{
RIKEN Center for Brain Science, Wako-shi, Japan, amari@brain.riken.jp}}
\date{}

\maketitle

\begin{abstract}
It is known that any target function is realized in a sufficiently small neighborhood of any randomly connected deep network, provided the width (the number of neurons in a layer) is sufficiently large.  There are sophisticated analytical theories and discussions concerning this striking fact, but rigorous theories are very complicated.  We give an elementary geometrical proof by using a simple model for the purpose of elucidating its structure.  We show that high-dimensional geometry plays a magical role: When we project a high-dimensional sphere of radius 1 to a low-dimensional subspace, the uniform distribution over the sphere shrinks to a Gaussian distribution with negligibly small variances and covariances.
\end{abstract}

\section{Introduction}

Deep learning has achieved great success in various applications by using very large networks.  However, it is only for a few years that theoretical foundations for sufficiently wide networks have been studied, where randomly connected initial weights play a fundamental role. \citet{JGH2018} showed that the optimal solution always lies sufficiently close to any networks with randomly assigned initial weights with high probability.  Hence, stochastic descent learning is described by a linear equation in the function space, describing learning behaviors of a neural network. This is a surprising result and there are many related theories (see, for example, \citet{LXSBNSP2019} and \citet{ADHLSW2019}). There are direct analytical theories explaining the fact that a target function exists in a small neighborhood of any randomly generated networks: See, for example, \citet{ALL2019} and \citet{BJTX2019}.

Unfortunately, these papers use mathematically rigorous analysis, so that it is not easy to follow and to understand the essence intuitively.  It is desirable to show a simple proof of this interesting fact to give insight for further developments.  The present paper tries to present an elementary proof that any target function lies in a small neighborhood of any randomly generated networks with high probability.

We consider an extremely high-dimensional sphere of radius 1 consisting of random parameter vectors.  Let their dimensions be $p$.  We project it to a low-dimensional subspace of dimension $n$, $n \ll p$, which corresponds to the number of training examples.  A uniform distribution over the high-dimensional sphere is projected to a Gaussian distribution of mean 0 and variance $1/p$, a sharply concentrated distribution. This geometrical fact explains that any true distribution exists in small neighborhoods of randomly generated networks.

Randomly connected networks are ubiquitous and have been studied in various situations. Early works, called statistical neurodynamics, are \citet{amari1971, amari1974, AYK1977} and \citet{rozonoer2969}, where macroscopic dynamics of randomly connected networks are studied. 
\citet*{SCS2015} showed that chaotic dynamics appears universally in a random network.  Another application is found in \citet{AATM2013}.

Statistical neurodynamics is applied to deep neural networks by \citet{PLRSG2016} to show how input signals are propagated through layers of networks, where the existence of the chaotic regime is found.  \citet{SGGS2016} applied the method to backward error propagation, showing the limit of applicability of error-backpropagation learning in the chaotic situation. There are many related works thereafter.  \citet{AKO2019a, AKO2019b} also studied the statistical neurodynamics of deep networks and elucidated the structure of Fisher information matrix in such a network.  \citet{KAA2019a, KAA2019b} developed a new direction of study by using the distribution of the eigenvalues of the Fisher information matrix of randomly connected networks. The present article is continuation along these lines of research.

\section{Deep neural networks}

We show a typical deep network, which has $d$-dimensional input ${\bm{x}}$ and scalar output $y$.  (It is easy to treat a network with multiple outputs in a similar way, which we do not do here.)  Its behavior is written as
\begin{equation}
    y = f({\bm{x}}, {\bm{\theta}}),
\end{equation}
where ${\bm{\theta}}$ is a vector parameter composed of connection weights and biases in all layers.  The network consists of $L$ layers. Each layer includes $p$ neurons, where $p$ is sufficiently large. (We may consider that layer $l$ includes $\alpha_l p$ neurons, but we assume $\alpha_l=1$ for simplicity).  The input to layer $l$ is the output of layer $l-1$, denoted by $\stackrel{l-1}{\bm{x}}$ (which is a $p$-dimensional vector) and the input-output relation of layer $l$ is
\begin{equation}
    \stackrel{l}{x_i} = \sum \varphi
    \left( {\bm{w}}_i \cdot
    \stackrel{l-1}{\bm{x}} \right),
\end{equation}
where $\stackrel{l}{x_i}$ is the $i$-th component of $\stackrel{l}{\bm{x}}$, $\varphi$ is an activation function, and $\stackrel{l}{\bm{w}_i}$ is a weight vector of the $i$-th neuron in layer $l$, whose components are denoted as $\stackrel{l}{w_{ij}}$.  We include the bias term in $\stackrel{l}{{\bm{w}}_i}$ for the sake of simplicity such that
\begin{eqnarray}
 \stackrel{l}{{\bm{w}}_i} &=&
  \left( \stackrel{l}{w_{i1}}, \cdots, \stackrel{l}{w_{ip}} \;;\; \stackrel{l}{w_{i0}} \right), \\
  \stackrel{l}{w_{i0}} &=& b_i
\end{eqnarray}
and add $\stackrel{l-1}{x_0}=1$ in $\stackrel{l-1}{\bm{x}}$ as its $0$-component. The final output $y$ comes from layer $L$ as
\begin{equation}
    y = f({\bm{x}}, {\bm{\theta}}) = 
    \sum_i v_i \stackrel{L}{x_i},
\end{equation}
assuming that the output is a linear function of $\stackrel{L}{\bm{x}}$, where ${\bm{v}}=\left(v_i \right)$ is the connection weight vector of the output neuron.  See Fig. 1.  We use the standard parameterization.  But we obtain the same result, even when we use the NTK parameterization \citep{LXSBNSP2019}.

\begin{figure}
    \centering
    \includegraphics[width=10cm]{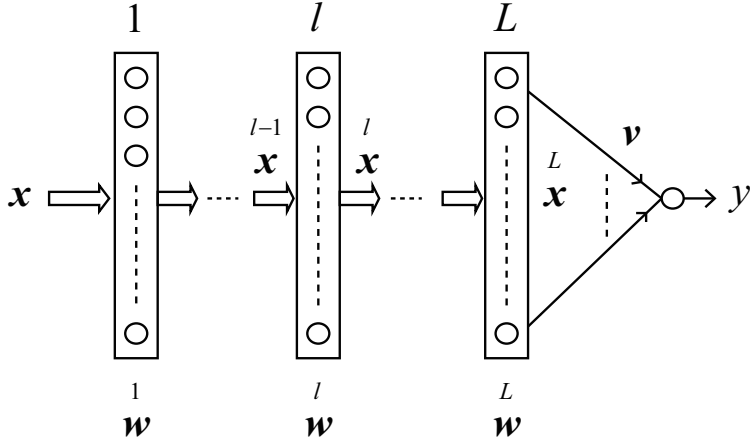}
    \caption{Deep Neural Networks}
    \label{fig:1}
\end{figure}

A random network is generated subject to the following probability law:
\begin{eqnarray}
  v_i &\sim& N \left(0, \frac{\sigma^2_v}{p} \right), \\
  \stackrel{l}{w_{ij}} &\sim& N 
  \left( 0, \frac{\stackrel{l}{\sigma^2_w}}p \right),
  \quad j \ne 0, \\
  \stackrel{l}{w_{io}} &\sim& N
  \left(0, \sigma^2_b \right), \quad 
   j=0,
\end{eqnarray}
independently, where $N \left(0, \sigma^2 \right)$ denotes the Gaussian distribution with mean $0$ and variance $\sigma^2$.  We may put $\sigma^2_v = \; \stackrel{l}{\sigma^2_w} \;=\; \stackrel{l}{\sigma^2_b} \;= 1$ in the following for the sake of simplicity, because we have interest mostly in qualitative structure, not in quantitative behaviors.

Let $D = \left\{ \left({\bm{x}}_1, y_1 \right), \cdots, \left( {\bm{x}}_n, y_n \right) \right\}$ be the set of $n$ training data.  We compose the $n$-dimensional output column vector
\begin{equation}
    {\bm{f}} = \left[ y_1, \cdots, y_n  
    \right]^T.
\end{equation}           
We also denote the output vector of the network of which parameters are ${\bm{\theta}}$ by
\begin{equation}
    {\bm{f}}_{\bm{\theta}} = 
    \left[ y \left({\bm{x}}_1, {\bm{\theta}} \right), \cdots y \left({\bm{x}}_n,
    {\bm{\theta}} \right) \right]^T.
\end{equation}

We assume that the outputs are bounded,
\begin{equation}
  \left| f_s \right| < c, \quad   
  s=1, \cdots, n
\end{equation}
for some constant $c$, so that
\begin{equation}
    \left\| {\bm{f}} \right\|_2
    < c \sqrt{n}.
\end{equation}
We put
\begin{equation}
    F = \left\{
     {\bm{f}} \left|
     \|{\bm{f}}\|_2 < c \sqrt{n}
     \right.
    \right\}.
\end{equation}
The output vector of a random network ${\bm{f}}_{\bm{\theta}}$ belongs to $F$ with high probability (whp), that is, with probability which tends to 1 as $p$ goes to infinity.  We further assume that inputs ${\bm{x}}_s$ ($s=1, \cdots, n$) in the training data are randomly and independently generated, and their components $\left|x_{si} \right|$ are bounded.

When we use the loss of squared errors, the optimal solution for minimizing the loss is given by
\begin{equation}
   {\bm{\theta}}^{\ast} = \mathop{\arg\min}_{\bm{\theta}}
   \sum^n_{s=1} \left\{
    y_s-f \left({\bm{x}}_s, {\bm{\theta}} 
    \right) \right\}^2.
\end{equation}

Under the above assumptions, we prove the following theorem.

\begin{theorem}\upshape
\label{theorem1}
The optimal solution ${\bm{\theta}}^{\ast}$ lies in a small neighborhood (the size of the neighborhood converging to $0$ as $p$ tends to infinity) of any ${\bm{\theta}}_0$, which is the parameters of any randomly generated network, with high probability.
\end{theorem}

\section{Linear theory for one hidden layer networks}

We consider a one hidden layer network in the beginning,
\begin{equation}
    f({\bm{x}}, {\bm{\theta}}) = \sum
    v_i \varphi \left( {\bm{w}}_i \cdot
    {\bm{x}} \right).
\end{equation}
We further fix randomly generated ${\bm{w}}_i$ (weights and biases).  Then, modifiable variables are only ${\bm{v}}= \left(v_1, \cdots, v_p \right)^T$ and the input-output relation is linear,
\begin{equation}
    {\bm{f}} = {\rm{\bf X}} {\bm{v}},
\end{equation}
where ${\rm{\bf X}}$ is $n \times p$ matrix, whose components are the outputs of the last layer,
\begin{equation}
   X_{si} = \varphi \left(
    {\bm{w}}_i \cdot {\bm{x}}_s \right), \quad
    s=1, \cdots, n \;;\; i=1, \cdots, p.
\end{equation}

This is a linear regression problem without noise, so we can easily analyze its behavior.  We give an elementary proof of Theorem \ref{theorem1} for this model.  Let ${\bm{v}}_0$ be a randomly generated parameter vector.  For the optimal parameters ${\bm{v}}^{\ast}$ satisfying ${\bm{f}}^{\ast} = {\rm{\bf X}}{\bm{v}}^{\ast}$, where ${\bm{f}}^{\ast} \in F$ is the teacher vector given from $D$, we put
\begin{equation}
    {\bm{v}}^{\ast} = {\bm{v}}_0 + \Delta
    {\bm{v}}.
\end{equation}
Then, we have 
\begin{equation}
    {\bm{f}}^{\ast} = {\rm{\bf X}}
    \left( {\bm{v}}_0 + \Delta{\bm{v}}
    \right).
\end{equation}
The output of the initial random neural network is
\begin{equation}
    {\bm{f}}_{\bm{v}} = {\rm{\bf X}}{\bm{{v}}}_0, 
\end{equation}
and the error vector is
\begin{equation}
   {\bm{e}} =  {\bm{f}}^{\ast} -
   {\bm{f}}_{\bm{v}}.
\end{equation}
So we have
\begin{equation}
 \label{eq:am2019122620}
    {\bm{e}} = {\rm{\bf X}} \Delta
    {\bm{{v}}}.
\end{equation}

The generalized inverse of ${\rm{\bf X}}$ is a $p \times n$ matrix defined by
\begin{equation}
    {\rm{\bf X}}^{\dagger} =
    {\rm{\bf X}}^T \left( {\rm{\bf X}}
    {\rm{\bf X}}^{T} \right)^{-1} =
    {\rm{\bf X}}^T {\rm{\bf K}}^{-1},
\end{equation}
where 
\begin{equation}
    {\rm{\bf K}} = {\rm{\bf X}}
    {\rm{\bf X}}^T
\end{equation}
is an $n \times n$ matrix called the Gram matrix.  It is the neural tangent kernel (Jacot et al., 2018) defined by
\begin{eqnarray}
  {\rm{\bf K}} &=& \left( K_{st} \right), \\
  K_{st} &=& \partial_{\bm{\theta}} f
  \left({\bm{x}}_s, {\bm{\theta}} \right)
  \cdot \partial_{\bm{\theta}} f
  \left({\bm{x}}_t, {\bm{\theta}} \right),
\end{eqnarray}
where $\cdot$ is the inner product.  The minimal norm solution of (\ref{eq:am2019122620}) is written as
\begin{equation}
    \Delta {\bm{v}}^{\ast} = {\rm{\bf X}}^T
    {\rm{\bf K}}^{-1}{\bm{e}},
\end{equation}
and the general solutions are
\begin{equation}
    \Delta {\bm{v}} =
    \Delta {\bm{v}}^{\ast} +{\bm{n}},
\end{equation}
where ${\bm{n}}$ is an arbitrary null vector belonging to the null subspace $N$ of ${\rm{\bf X}}$,
\begin{equation}
    N = \left\{ {\bm{n}} \left| \; 
    {\rm{\bf X}} {\bm{n}} = 0
    \right.  \right\}.
\end{equation}
 
We study the kernel matrix ${\rm{\bf K}}$ for evaluating the magnitude of $\Delta {\bm{v}}^{\ast}$.  Let us put
\begin{equation}
    u_{si} = {\bm{w}}_i \cdot {\bm{x}}_s, \quad
    u_{ti} = {\bm{w}}_i \cdot {\bm{x}}_t.
\end{equation}
Then, they are jointly Gaussian with mean $0$ and $n$ pairs of variables $\left( u_{si}, u_{ti} \right), i=1, \cdots, n$, are independent subject to the same probability distribution.  
Their correlation 
\begin{equation}
 \label{eq:am2820191226}
    \sigma^2_{st} = {\rm{E}} \left[
     u_{si} u_{ti} \right] =
     \sigma^2_w {\bm{x}}_s \cdot 
     {\bm{x}}_t
     + \sigma^2_b
\end{equation}
depends on the inner product of two inputs ${\bm{x}}_s$ and ${\bm{x}}_t$ and is the same for any $i$.  The components of ${\rm{\bf K}}$ 
\begin{equation}
 \label{eq:am3220200114}
    K_{st} = \sum_i \varphi \left(u_{si}
    \right) \varphi \left( u_{ti} \right)
\end{equation}
are sums of $p$ iid variables.
Because of the law of large numbers, we have whp
\begin{equation}
 \label{eq:am3020191226}
  \frac 1p K_{st} =  {\rm{E}} \left[
    \varphi \left(u_{si} \right) \varphi
    \left( u_{ti} \right)
   \right].  
\end{equation}
Hence ${\rm{\bf K}}^{-1}$ is of order $1/p$.  The explicit form of (\ref{eq:am3020191226}) is given in \citet{LXSBNSP2019} and \citet{AKO2019a, AKO2019b} for the sigmoid function and for ReLU function.   

This proves that any component of $\Delta v^{\ast}_i$ is of order $1/p$, provided $n$ is fixed and $p \gg n$. Hence, its $L_2$ norm is
\begin{equation}
    \| \Delta {\bm{v}}^{\ast}\|_2 = O
    \left( \frac 1{\sqrt{p}} \right),
\end{equation}
whereas the $L_2$ norm of the initial random ${\bm{v}}_0$ is not small,
\begin{equation}
    \|{\bm{v}}_0\|_2 =
    \sigma^2_v = O(1).
\end{equation}
This shows that a true solution ${\bm{v}}_0+ \Delta{\bm{v}}^{\ast}$ exist in a $(1/ \sqrt{p})$-neighborhood of any random weight vector ${\bm{v}}_0 \; \mbox{whp}$.

It should be verified that the minimum norm solution $\Delta {\bm{v}}^{\ast}$ is obtained by learning.  $\Delta {\bm{v}}^{\ast}$ is in proportion to the gradient vector of the loss, 
\begin{equation}
    \Delta {\bm{v}}^{\ast} 
    \propto \sum_s
    \partial_{\bm{v}} f({\bm{x}}_s, {\bm{\theta}}) e_s =
    {\rm{\bf X}}^T {\bm{e}},
\end{equation}
where ${\bm{e}}= \left( e_s \right)$.
Hence,
\begin{equation}
    {\bm{n}}\cdot \Delta {\bm{v}}^{\ast}  \propto \left({\rm{\bf X}}{\bm{n}}
    \right)^T {\bm{e}}
    =0. 
\end{equation}
$\Delta{\bm{v}}^{\ast}$ is orthogonal to the null subspace $N$, not including the null components.

{\textbf{Remark 1.}}  Mini batch stochastic gradient learning does not change the components of the null direction included in the initial ${\bm{v}}_0$. Hence, the minimal solution itself is not derived by learning.  When we add a decay term in the learning equation, we have the minimal solution.  The minimal solution would have the smallest generalization error. 

{\textbf{Remark 2.}}  When $n$ is large, we can show
\begin{equation}
    \Delta v_i = O \left(
    \frac {n^2}p
    \right).
\end{equation}
Hence, our theory does not hold for large $n$, $n>O(\sqrt{p})$.  

{\textbf{Remark 3.}}
When the target function $f(\bm{x})$ is sufficiently smooth, as for example band limited, $\Delta v_i=O \left(\frac 1p \right)$ would be possible for approximating $f({\bm{x}})$ within a small error $\varepsilon$.  We need a separate proof for this along the line of \citet{BJTX2019}.

{\textbf{Remark 4.}}  We have not discussed the generalization error.  However, our situation of random ${\rm{\bf X}}$ would guarantee the ``non-prescient'' situation of double descent shown in \citet{BHX2019}.

\section{Geometric perspective}

The randomly chosen vectors ${\bm{v}}_0$ are distributed in the space of parameters. Because of the law of large numbers, we have
\begin{equation}
    \|{\bm{v}}_0 \|^2 = \sigma^2_v
   + O \left( \frac 1p \right).
\end{equation}
That is, the $L_2$ norm of random ${\bm{v}}_0$ is almost equal to $\sigma^2_v$.  We put
\begin{equation}
    \sigma^2_v = 1
\end{equation}
for simplicity. Then random ${\bm{v}}_0$’s are uniformly distributed on the sphere $V$ of radius 1 of which center is the origin, having infinitesimal small deviations in the radial directions.  We project these ${\bm{v}}_0$’s to the $n$-dimensional subspace $S$ orthogonal to $N$.  $S$ consists of the minimum norm solutions ${\bm{v}}^{\ast}$ of
\begin{equation}
    {\bm{f}}^{\ast} = 
    {\rm{\bf X}} {\bm{v}}^{\ast}
\end{equation}
for all ${\bm{f}}^{\ast} \in F$.
We prove the following theorem, showing that the uniform distribution over $V$ shrinks to a Gaussian distribution with variance $1/p$.

\begin{theorem}\upshape
The projection of the uniform distribution over $V$ to $S$ gives asymptotically a Gaussian distribution with mean 0 and covariance matrix $(1/p)$ $I$, where $I$ is the identity matrix.
\end{theorem}

\begin{proof}
We first give an intuitive explanation of the projection. We consider a $(p \;–\; 1)$-dimensional sphere $V$ in ${\bm{R}}^p$ and project it to a 1-dimensional line $S$.  Let $z$ be a position on the line.  Then, the inverse image of the projection of a small line element $[z, z + dz]$ is a slice of the sphere orthogonal to the line, which is a $(p\;–\;2)$-dimensional sphere with width $dz$ and its radius is $\sqrt{1-z^2}$. See Fig. 2. 
\begin{figure}
    \centering
    \includegraphics[width=12cm]{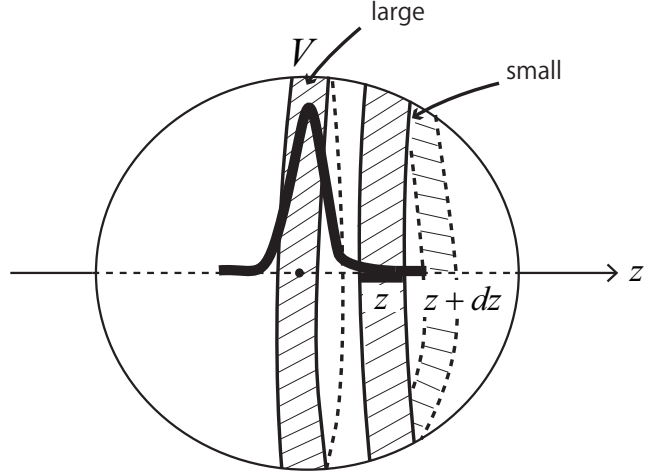}
    \caption{Projection of $V$ to $Z$-axis}
    \label{fig:2}
\end{figure}
The total mass of the sliced sphere is
\begin{equation}
    \left( \sqrt{1-z^2} 
    \right)^{p-2}
    c(z)dz,
\end{equation}
where the term $c(z)$ is added to be responsible for the inclination of the sliced sphere and the total volume of the sphere of radius 1.  This converges to 0 when $p$ is large except for the parts corresponding to $z \approx 0$.  This shows that the projection of the $(p\;–\;1)$-dimensional sphere to the line is concentrated around the origin. 
\end{proof}

We give a formal proof for the $n$-dimensional case.  Let $R$ be a point in the $n$-dimensional $S$, and its radius be $z$.  Then, the inverse image of the projection is a $(p\;–\;n-1)$-dimensional sphere sliced at $R$ (see Fig. 3), and its radius is $\sqrt{1-z^2}$.  Hence, the density $q(z)$ of projected mass is calculated as
\begin{figure}
    \centering
    \includegraphics[width=12cm]{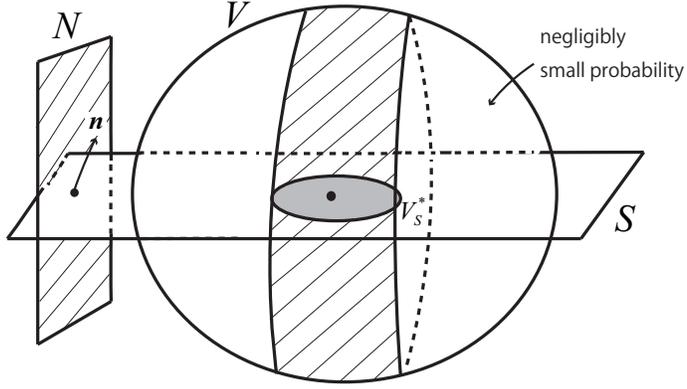}
    \caption{Inverse projection of $V^{\ast}_S$ covers most part of $V$}
    \label{fig:3}
\end{figure}
\begin{equation}
    q(z) = c (z) \left(1-z^2 \right)^{\frac{p-n-1}{2}}.
\end{equation}
We put
\begin{equation}
    z^2 = \frac {\varepsilon}p,
\end{equation}
and then
\begin{equation}
    q(z)
     = c \left\{ 1-\frac{\varepsilon}p \right\}^{\frac p{\varepsilon} \frac{\varepsilon}2}
     = c \exp \left\{ -\frac{\varepsilon}2
     \right\}.
\end{equation}
Since we assume $p \gg n$, this gives the Gaussian distribution 
\begin{equation}
    q(z) = c \exp \left\{
     -\frac{pz^2}2
    \right\}.
\end{equation}
proving that almost all random ${\bm{v}}$ are mapped inside a small sphere in $S$ of radius $1/\sqrt{p}$, or more precisely $1/\sqrt{p-n}$.

For any ${\bm{f}}^{\ast}$, the minimum norm solution ${\bm{v}}^{\ast}$ of ${\bm{f}}^{\ast} = {\rm{\bf X}}{\bm{v}}^{\ast}$,
\begin{equation}
    {\bm{v}}^{\ast} = {\rm{\bf X}}
    {\rm{\bf K}}^{-1}{\bm{f}}^{\ast}
\end{equation}
sits in the $(1/p)$-neighborhood of the origin in the $n$-dimensional $S$.  Its inverse image ${\bm{v}}^{\ast}+{\bm{n}}$ ${\bm{n}} \in N$ covers almost all parts of $V$ for any ${\bm{f}}^{\ast}$.  This is a geometrical explanation that the true solution lies in a neighborhood of any random ${\bm{v}}_0$.

To summarize, Theorem \ref{theorem1} is based on the following two asymptotic facts (see Fig. 3):
\begin{enumerate}
    \item[1)] The minimum optimum solution ${\bm{v}}^{\ast}$ lies in $(1/p)$-neighborhood $V^{\ast}_S$ of the origin of $S$ for all ${\bm{f}}^{\ast} \in F$.
   \item[2)] The inverse image of the neighborhood $V^{\ast}_S$, that is $\left\{ {\bm{v}}^{\ast} + {\bm{n}} \left| {\bm{f}}^{\ast} \in F, {\bm{n}} \in N \right. \right\}$ covers almost all parts of $V \subset {\bm{R}}^p$.
\end{enumerate}

\section{Analysis of general networks with one hidden layer}

The connections ${\bm{w}}_i$ were randomly generated and fixed in the previous linear model. We consider here the case where both ${\bm{v}}$ and ${\bm{w}}_i$ are modifiable.  Small deviations of ${\bm{v}}$ and ${\bm{w}}_i$ to ${\bm{v}}+\Delta {\bm{v}}$ and ${\bm{w}}_i+ \Delta {\bm{w}}_i$ give a change of function $f({\bm{x}}, {\bm{\theta}})$ as
\begin{equation}
    \Delta f \left( {\bm{x}}, {\bm{v}}, {\bm{w}}_i \right)
    = \sum \frac{\partial f}{\partial v_i}
    \Delta v_i + \sum 
    \frac{\partial f}{\partial w_{ij}}
    \Delta w_{ij}.
\end{equation}
We define ${\rm{\bf {X}}}^{(1)}$ and ${\rm{\bf {X}}}^{(2)}$ by
\begin{eqnarray}
   X^{(1)}_{si} &=&
  \frac{\partial f \left( {\bm{x}}_s \right)}{\partial v_i}, \\
   X^{(2)}_{sij} &=&
  \frac{\partial f \left( {\bm{x}}_s \right)}{\partial w_{ij}}.
\end{eqnarray}
where ${\rm{\bf X}}^{(1)}$ is the same as the previous ${\rm{\bf X}}$ and
\begin{equation}
  {\rm{\bf X}}^{(2)}_{sij} = v_i \varphi'
  \left(u_{si} \right) x_{sj}.
\end{equation}
We use a vectorized $d{\bm{\theta}} = \left( \Delta v_i, \Delta w_{ij} \right)^T$ for denoting small deviations $\Delta {\bm{v}}$ and $\Delta {\bm{w}}_i$, and use index $I$ for denoting the components of $d{\bm{\theta}} = \left( d \theta_I \right)$.  The extended ${\rm{\bf X}}$ is
\begin{equation}
 \label{eq:am4220191226}
  {\rm{\bf X}} = \left[
   {\rm{\bf X}}^{(1)}, {\rm{\bf X}}^{(2)}
  \right]  
\end{equation}
and
\begin{equation}
    d{\bm{f}} = {\rm{\bf X}} \Delta {\bm{\theta}}.
\end{equation}
  
For error vector
\begin{equation}
    {\bm{e}} = {\bm{f}}^{\ast} - {\bm{f}}_{\bm{\theta}}, 
\end{equation}
the equation to determine the deviation
\begin{equation}
    \Delta {\bm{\theta}} = 
    \left[
     \Delta {\bm{v}}, \Delta {\bm{w}}_i
    \right]^T
\end{equation}
for the optimal solution ${\bm{\theta}}^{\ast} = {\bm{\theta}} + \Delta{\bm{\theta}}$ is written as 
\begin{equation}
    \Delta {\bm{\theta}} = 
    {\rm{\bf X}}^{\dagger} {\bm{e}} =
    {\rm{\bf X}}^T {\rm{\bf K}}^{-1}{\bm{e}},
\end{equation}
provided $\Delta{\bm{\theta}}$ is small. We evaluate the tangent kernel ${\rm{\bf K}}={\rm{\bf X}}{\rm{\bf X}}^T$ in the present case.

From (\ref{eq:am4220191226}), ${\rm{\bf K}}$ is decomposed as
\begin{equation}
  {\rm{\bf K}} = {\rm{\bf K}}^{(1)} +
  {\rm{\bf K}}^{(2)},
\end{equation}
where
\begin{equation}
   {\rm{\bf K}}^{(i)} =  
   {\rm{\bf X}}^{(i)}{\rm{\bf X}}^{(i)T}, \quad i= 1, 2.
\end{equation}
${\rm{\bf K}}^{(1)}$ is the same as the previous one (\ref{eq:am3220200114}) and ${\rm{\bf K}}^{(2)}$ is written in the component form as
\begin{equation}
   K^{(2)}_{st} =
   \sum_{i, j} v^2_i \varphi'
   \left(u_{si} \right) \varphi'
   \left(u_{ti} \right) x_{sj} x_{tj}.
\end{equation}

In order to evaluate $K^{(2)}_{st}$, we remark that $p$ pairs of variables $\left(u_{si}, u_{ti}\right), i=1, \cdots, p$, are iid (identically and independently distributed) random variables. So the law of large numbers guarantees that it converges to
\begin{equation}
   \frac 1p K^{(2)}_{st} =  E \left[ \varphi' \left(u_{si}\right) \varphi' \left(u_{ti} \right)\right]{\bm{x}}_s \cdot {\bm{x}}_t.
\end{equation}

We put
\begin{equation}
    \chi^1_{st} = E \left[
     \varphi' \left(u_{si} \right)
     \varphi' \left(u_{ti} \right)
    \right],
\end{equation}
of which exact form depends on $\varphi$ and it is calculated explicitly for various activation functions \citep[see e.g.,][]{LXSBNSP2019, AKO2019a, AKO2019b}.  Then, 
\begin{equation}
    K^{(2)}_{st} =
    p \chi^1_{st}{\bm{x}}_s \cdot
    {\bm{x}}_t.
\end{equation}

From this, we see that
\begin{equation}
  {\rm{\bf K}} = O(p), \quad {\rm{\bf K}}^{-1}
  = O \left( \frac 1p \right).
\end{equation}
This guarantees that $\Delta \theta_I$ is small,
\begin{equation}
    \Delta \theta_I = O \left( \frac 1p \right),
\end{equation}
proving Theorem \ref{theorem1} in the present case.

\section{General deep networks}

For a deep network with $L$ layers, the output functions is nested as,
\begin{eqnarray}
   f({\bm{x}}, {\bm{\theta}}) &=& \sum v_i
   \varphi \left( \stackrel{L}{\bm{w}_i} \cdot \stackrel{L-1}{\bm{x}} \right), \\
   \stackrel{l}{x_i} &=& \varphi
   \left( \stackrel{l}{\bm{w}_i} \cdot 
   \stackrel{l-1}{\bm{x}} \right), \quad
   l=1, \cdots, L-1.
\end{eqnarray}
    
A small deviation of $f$ due to small deviation $\Delta {\bm{\theta}}$ of ${\bm{\theta}}=\left({\bm{v}}, \stackrel{L}{\bm{w}_i}, \cdots, \stackrel{1}{\bm{w}_i} \right)$ is
\begin{equation}
    \Delta f = \sum \varphi
    \left( \stackrel{L}{\bm{w}}_i \cdot {\bm{x}} \right) \Delta
    v_i + \sum_{l, i} 
    \frac{\partial f}{\partial \stackrel{l}{{\bm{w}}_i}}
    \Delta \stackrel{l}{{\bm{w}}_i}.
\end{equation}
For vector ${\bm{f}}= \left( f \left( {\bm{x}}_s, {\bm{\theta}} \right) \right)$, $s=1, \cdots, n$, this is written as
\begin{equation}
    \Delta {\bm{f}} = {\rm{\bf X}} \Delta
    {\bm{\theta}},
\end{equation}
where
\begin{eqnarray}
  {\rm{\bf X}} &=& \left(
   \frac{\partial {\bm{f}}}{\partial {\bm{\theta}}}
  \right) = \left[
     \stackrel{L+1}{\rm{\bf X}}, 
     \stackrel{L}{\rm{\bf X}}, \cdots,
     \stackrel{1}{\rm{\bf X}}
   \right], \\
  \stackrel{l}{{\rm{\bf X}}_i} &=&
  \left(
    \frac{\partial {\bm{f}}}{\partial \stackrel{l}{ {\bm{w}}_i}}
  \right), \quad
  \stackrel{L+1}{\rm{\bf X}} =
  {\rm{\bf X}}^{(1)}.
\end{eqnarray}

From $n$ training examples, we have a similar equation
\begin{eqnarray}
    \Delta {\bm{\theta}} &=& {\rm{\bf X}}^{\dagger}
    {\bm{e}} =
    {\rm{\bf X}}^T {\rm{\bf K}}^{-1}
     {\bm{e}} \\
    {\rm{\bf K}} &=& {\rm{\bf X}}{\rm{\bf X}}^T
    = \sum^{L+1}_{l=1} \stackrel{l}{\rm{\bf X}}
    \stackrel{l}{\rm{\bf X}^T}
\end{eqnarray}
for calculating the optimal $\Delta {\bm{\theta}}$.  ${\rm{\bf K}}$ is decomposed as
\begin{equation}
    {\rm{\bf K}} = \sum^{L+1}_{l=1}
    \stackrel{l}{\rm{\bf K}}.
\end{equation} 
The tangent kernel ${\rm{\bf K}}$ and $\stackrel{l}{\rm{\bf{K}}}$ are calculated recursively by \citet{JGH2018} and \citet{LXSBNSP2019}.  See Appendix.  Since $\stackrel{l}{\rm{\bf K}}$ is of order $p$, we finally have
\begin{equation}
  {\rm{\bf K}} = O(L p)  
\end{equation}
and
\begin{equation}
  {\rm{\bf K}}^{-1} = O \left( \frac 1{Lp} \right),  
\end{equation}
proving the theorem.

\section*{Conclusions}

Randomly connected over-parameterized wide networks have magical power that any target functions are found in small neighborhoods of any random networks. By using simple models, we have elucidated this situation, giving an elementary proof of it. This can be illustrated from high dimensional geometry: The projection of the uniform distribution over a high-dimensional unit sphere to a low-dimensional subspace gives a Gaussian distribution which is sharply concentrated in a small neighborhood of the origin.

The present paper makes it clear that empirical finite samples can be well approximated near random initialization. Its magic lies in the null directions $N$ existing in the space of parameters. It elucidates geometrical picture of the equivalence relation of functions originated from the finiteness of samples.  We remark that the null space $N$ exists even when $n \rightarrow \infty$, when the neurons are continuously arranged to form a neural field.  See \citealp{SIIHSMM2018}.

The author believes that the present geometrical method is useful not only explaining the surprising power of random initialization for training, but also useful for elucidating the generalization errors and its relation to the neural tangent kernel. It would be possible to explain the power of NTRF (neural tangent random feature) from geometry.  see, e.g., \citealt*{CG2019}.

\section*{Appendix:
Calculations of ${\rm{\bf K}}$}

We follow \citet{LXSBNSP2019} for calculations of ${\rm{\bf K}}$, slightly changing notations.  We put
\begin{eqnarray}
   \stackrel{l}{J}_{i_l j_{l-1}} &=&
   \frac{\partial}{\partial \stackrel{l-1}{x_{j_l}}}
   \varphi \left( \stackrel{l}{{\bm{w}}} \cdot
   \stackrel{l-1}{{\bm{x}}}\right)
   = \varphi' \left( \stackrel{l}{u_{i_l}} \right)
   \stackrel{l}{w_{i_l j_{l-1}}}, \\
   \stackrel{l}{\bm{J}}_{ij_l} &=& \sum
   \stackrel{L}{J}_{i j_{L-1}}
   \stackrel{L-1}{J_{j_{L-1}}}{}_{j_{L-2}}
   \cdots 
   \stackrel{l+1}{J}_{j_{l+1}j_l}.
\end{eqnarray}
Then,
\begin{equation}
    \stackrel{l}{X_{si}} = \sum v_i
    \stackrel{l+1}{\bm{J}}_{ij_l}
    \varphi' \left( \stackrel{l}{u_{sj_l}} \right)
    \stackrel{l-1}{x}_{sj_l}.
\end{equation}

We evaluate the layer $l$ part of the neural tangent kernel
\begin{equation}
    \stackrel{l}{K}_{st} =
    \sum_{i, l} v^2_i \stackrel{l+1}{\bm{J}}_s
    \stackrel{l+1}{\bm{J}}_t \varphi'
    \left( \stackrel{l}{u}_s \right)
    \varphi'
    \left( \stackrel{l}{u}_t \right)
    \stackrel{l-1}{\bm{x}}_s \cdot
     \stackrel{l-1}{\bm{x}}_t, 
\end{equation}
where some suffices are omitted.  We use the mean field paradigm to assume that the law of large numbers holds.  Moreover, we assume that $\varphi' \left(\stackrel{l}{\bm{w}}_i \cdot {\bm{x}} \right)$ has self-averaging property.  We then have
\begin{equation}
    \stackrel{l}{K}_{st} =
    p {\rm{E}} \left[ \stackrel{l+1}{\bm{J}}_s
    \stackrel{l+1}{\bm{J}}_t \right]
    E \left[ \varphi' 
    \left( \stackrel{l}{u}_s \right)
    \varphi' \left( \stackrel{l}{u_t}\right)\right]
    {\rm{E}} \left[ \stackrel{l-1}{\bm{x}}_s \cdot
    \stackrel{l-1}{\bm{x}}_t
    \right].
\end{equation}
The quantities 
\begin{equation}
    \stackrel{l}{\chi^1_{st}} =
    {\rm{E}}
    \left[
     \varphi' \left( \stackrel{l}{u_s} \right)
     \varphi' \left( \stackrel{l}{u_t}\right)
    \right]
\end{equation}
can recursively be calculated.

We put
\begin{eqnarray}
    \stackrel{l}{\chi^0_{st}} &=&
    {\rm{E}} \left[
     \varphi \left( \stackrel{l}{u_{si}} \right) \varphi
     \left(
      \stackrel{l}{u_{ti}}
     \right)
    \right] = {\rm{E}}
    \left[
     \stackrel{l}{x_{si}}
     \stackrel{l}{x_{ti}}
    \right] \\
    \stackrel{l}{\chi^1_{st}} &=&
    {\rm{E}} \left[
     \varphi' \left( \stackrel{l}{u_{si}} \right) \varphi'
     \left(
      \stackrel{l}{u_{ti}}
     \right)
    \right].
\end{eqnarray}
Then, the following recursive equations hold for the tangent kernels $\stackrel{l}{K_{st}}$,
\begin{equation}
  \stackrel{l}{K_{st}} = 
  \sigma^2_{w} \left(
   \stackrel{l-1}{K_{st}} \;
   \stackrel{l-1}{\chi^1_{st}}
   + \stackrel{l-1}{\chi^0_{st}}
  \right) + \sigma^2_b,
\end{equation}
with
\begin{equation}
 \stackrel{1}{K_{st}} = 
 \sigma^2_{w} {\bm{x}}_s \cdot
 {\bm{x}}_t + \sigma^2_b.
\end{equation}


\begin{thebibliography}{999}
%
\bibitem[Allen-Zhu et al.(2019) Allen-Zhu, Li and Liang]{ALL2019} Allen-Zhu, Z., Li, Y., Liang, Y., Learning and generalization in overparameterized neural networks, going beyond two layers. arXiv:1811.04918v2, (2019).
%
\bibitem[Amari(1971)]{amari1971} Amari, S., Characteristics of randomly connected threshold-element networks and network systems. Proc. IEEE., 59, 1, 35--47, (1971).
%
\bibitem[Amari(1974)]{amari1974} Amari, S., A method of statistical neurodynamics. Kybernetik, 14, 201--215, (1974).
%
\bibitem[Amari et al.(2013) Amari, Ando, Toyoizumi and Masuda]{AATM2013} Amari, S., Ando, H., Toyoizumi, T., Masuda, N., State concentration exponent as a measure of quickness in Kauffman-type networks, Physical Review, E 87, 022814, (2013).
%
\bibitem[Amari et al.(2019a) Amari,  Karakida and Oizumi]{AKO2019a} Amari, S.,  Karakida, R., Oizumi, M., Fisher information and natural gradient learning in random deep networks. AISTATS 2019, arXiv:, (2019a).
%
\bibitem[Amari et al.(2019b) Amari,  Karakida and Oizumi]{AKO2019b} Amari, S., Karakida, R., Oizumi, M., Statistical neurodynamics of deep networks: Geometry of signal spaces. Nonlinear Theory and Its Applications, IEICE, vol.2, pp.1101--1115, (2019b).
%
\bibitem[Amari et al.(1977) Amari, Yoshida and Kanatani]{AYK1977} Amari, S., Yoshida, K., Kanatani, K., A Mathematical foundation for statistical neurodynamics. SIAM J. Appl. Math., 33, 95--126, (1977).
%
\bibitem[Arora et al.(2019) Arora, Du, Hu, Li, Salakhutdinov and Wang]{ADHLSW2019} Arora, S., Du, S.S., Hu, W., Li, Z., Salakhutdinov, R., Wang, R., On exact computation with an infinitely wide net. arXiv:1904.11955v1 (2019).

%
\bibitem[Bailey et al.(2019) Bailey, Ji,  Telgarsky and Xian]{BJTX2019} Bailey, B., Ji, Z., Telgarsky, M., Xian, R., Approximation power of random neural networks. arXiv:1906.07709v1, (2019).
%
\bibitem[Belkin et al.(2019) Belkin, Hsu and Xu]{BHX2019} Belkin, M., Hsu, D., Xu, J., Two models of double descent for weak feature. arXiv:1903.07571v1, (2019).

%
\bibitem[Cai et al.(2019) Cai, Gao, Hou, Chen, Wong, He, Zhang and Wang]{CGHCWHZW2019} Cai, T.,  Gao, R., Hou, J., Chen, S., Wong, D., He, D., Zhang, Z., Wang, L., A Gram-Gauss-Newton method learning overparametrized deep neural networks for regression problems. arXiv:1905.11675v1, (2019).
%
\bibitem[Cao et al.(2019) Cao and Gu]{CG2019} Cao, Y., Gu, Q., Generalization bounds of stochastic gradient descent for wide and deep neural networks. arXiv:1905.13210v3, NeurIPS, 2019.
%
\bibitem[Jacot et al.(2018) Jacot, Gabriel and Hongler]{JGH2018} Jacot, A., Gabriel, F.,   Hongler, C., Neural tangent kernel: Convergence and generalization in neural networks. Neurips 2018, arXiv:1806.07572v3, (2018).
%
\bibitem[Karakida et al.(2019a) Karakida, Akaho and Amari]{KAA2019a} Karakida, R., Akaho, S., Amari, S., Universal statistics of Fisher information in deep neural networks: Mean field approach. AISTATS, 2019; arXiv:1806.01316v2, (2019a).
%
\bibitem[Karakida et al.(2019b) Karakida, Akaho and Amari]{KAA2019b} Karakida, R., Akaho, S., Amari, S., The normalization method for alleviating pathological sharpness in wide neural networks. 33rd Conference on Neural Information Processing Systems (NeurIPS 2019), Vancouver, Canada, arXiv:1906.02926, (2019b).
%
\bibitem[Lee et al.(2019) Lee, Xiao, Schoenholz, Bahri, Novak, Sohl-Dickstein and J. Pennington]{LXSBNSP2019} Lee, J., Xiao, L.,  Schoenholz, A. S., Bahri, Y., Novak, R.,  Sohl-Dickstein, J.,  Pennington, J., Wide neural networks of any depth evolves as linear models under gradient descent.  arXiv:1902.06720v3, (2019).
%
\bibitem[Poole et al.(2016) Poole, Lahiri, Raghu, Sohl-Dickstein and Ganguli]{PLRSG2016} Poole, B., Lahiri, S., Raghu, M., Sohl-Dickstein, J., Ganguli, S., Exponential expressivity in deep neural network through transient chaos. in Advances in Neural Information Processing (NIPS), 3360--3368, (2016).
%
\bibitem[Rozonoer(1969)]{rozonoer2969} Rozonoer, L.I., Random logical nets, I, II, III. Avtomatika I Telemekhanica, nos. 5, 6, 7, 137--147, 99--109, 127--136, (1969).
%
\bibitem[Schoenholz et al.(2016) Schoenholz, Gilmer, Ganguli and Sohl-Dicksein]{SGGS2016} Schoenholz, S.S., Gilmer, J,, Ganguli, S., Sohl-Dickstein, J., Deep information propagation. ICLR'2017, arXiv: 1611.01232, (2016).
%
\bibitem[Sonoda et al.(2018) Sonoda, Ishikawa, Ikeda, Hagihara, Sawano, Matsubara and Murata]{SIIHSMM2018} Sonoda, S., Ishikawa, I.,  Ikeda, M., Hagihara, K., Sawano, Y., Matsubara, T.,  Murata, N., The global optimum of shallow neural network is attained by ridgelet transform. arXiv:1805.07517, (2018).
%
\bibitem
[Sompolinsky et al.(2015) Sompolinsky, Cristanti and Sommers]
{SCS2015} Sompolinsky, H.,  Cristanti, A., Sommers, H. J., Chaos in random neural networks. Physical Review, E91, 032802, (2015).
%
\bibitem[Yang (2019)]{yang2019} Yang, G., Scaling limits of wide neural networks with weight sharing: Gaussian process behavior, gradient independence, and natural tangent kernel derivation. arXiv:1902.04760, (2019).

%
\bibitem[Zhang et al.(2019) Zhang,  Martens and Grosse]{ZMG2019} Zhang, G.,  Martens, J., Grosse, R., Fast convergence of natural gradient descent for overparametrized neural networks. arXiv:1905.10961v1, (2019).


\end{thebibliography}
\end{document}